\documentclass{article}
\usepackage{dcase2021,amsmath,graphicx,url,times,booktabs, tabularx}

\usepackage[natbib, bibencoding=utf8, citestyle=numeric, bibstyle=ieee, maxbibnames=999, maxcitenames=2, mincitenames=1]{biblatex}
\usepackage{makecell}
\usepackage{booktabs,subcaption}
\usepackage{amsmath,amsfonts} % For pmb and mathbb
\usepackage[hidelinks, pdfusetitle]{hyperref}
\usepackage[capitalise, nameinlink, noabbrev]{cleveref}

\usepackage[acronym, shortcuts, nohypertypes={acronym}]{glossaries}
\newacronym{AI}{AI}{artificial intelligence}
\newacronym{ASC}{ASC}{acoustic scene classification}
\newacronym{CNN}{CNN}{convolutional neural network}
\newacronym{DL}{DL}{deep learning}
\newacronym{DNN}{DNN}{deep neural network}
\newacronym{FFNN}{FFNN}{feed-forward neural network}
\newacronym{TDNN}{TDNN}{time-delay neural network}
\newacronym{ML}{ML}{machine learning}
\newacronym{SGD}{SGD}{stochastic gradient descent}
\newacronym{ReLU}{ReLU}{rectified linear unit}

\title{
    Fairness and underspecification in acoustic scene classification:\\
    The case for disaggregated evaluations
}

\name{Andreas Triantafyllopoulos$^{1,2}$,
      Manuel Milling$^{2}$,
      Konstantinos Drossos$^3$,
      Bj\"orn W. Schuller$^{1,2,4}$
      }
\address{$^1$audEERING GmbH, Gilching, Germany\\
        $^2$EIHW -- Chair of Embedded Intelligence for Health Care and Wellbeing, University of Augsburg, Germany\\
        $^3$Audio Research Group, Tampere University, Tampere, Finland\\
        $^4$GLAM -- Group for Audio, Language, \& Music, Imperial College, London, UK\\
        atriant@audeering.com
 }

\begin{document}

\ninept
\maketitle

\begin{sloppy}

\begin{abstract}
\emph{Underspecification} and \emph{fairness} in \ac{ML} applications have recently become two prominent issues in the \ac{ML} community.
\Ac{ASC} applications have so far remained unaffected by this discussion, but are now becoming increasingly used in real-world systems where fairness and reliability are critical aspects.
In this work, we argue for the need of a more holistic evaluation process for \ac{ASC} models through \emph{disaggregated evaluations}.
This entails taking into account performance differences across several factors, such as city, location, and recording device.
Although these factors play a well-understood role in the performance of \ac{ASC} models, most works report single evaluation metrics taking into account all different strata of a particular dataset.
We argue that metrics computed on specific sub-populations of the underlying data contain valuable information about the expected real-world behaviour of proposed systems, and their reporting could improve the transparency and trustability of such systems.
We demonstrate the effectiveness of the proposed evaluation process in uncovering underspecification and fairness problems exhibited by several standard \ac{ML} architectures when trained on two widely-used \ac{ASC} datasets.
Our evaluation shows that all examined architectures exhibit large biases across all factors taken into consideration, and in particular with respect to the recording location.
Additionally, different architectures exhibit different biases even though they are trained with the same experimental configurations.
\end{abstract}


\glsresetall

\section{Introduction}
\label{sec:intro}

\Ac{ASC} has been established as a central task of artificial auditory intelligence, as exemplified by its prominent place in the DCASE challenge and workshop series~\citep{DCASE2017challenge, Mesaros2018_TASLP} and a generally broad accumulation of literature~\citep{Liu1998, barchiesi2015acoustic, Rakotomamonjy2017, Kun2017, ren2018deep}.
Overall, model performance has substantially improved through the years, and datasets have accordingly evolved to accommodate new challenges by incorporating factors shown to impact model performance.
For example, the exact geographical location of the recordings was identified as an important factor early on, with datasets accordingly adapted by keeping data from the same location in the same partitions~\citep{DCASE2017challenge, Mesaros2018_TASLP}.
The TUT Urban Acoustic Scenes 2018 Mobile dataset additionally introduced the recording device as a separate factor~\citep{Mesaros2018_DCASE}, with the development set consisting of multiple recording devices, and the evaluation set including an extra, unseen device.
Finally, the TAU Urban Acoustic Scenes 2019 dataset highlighted the importance that the city of origin has by introducing data from two additional cities in the evaluation set~\citep{Mesaros2018_DCASE}.

In general, the community is aware of the influence that recording devices and location have on model performance~\citep{Mesaros2019, heittola2020acoustic}.
Most works approach these factors from the perspective of \emph{domain mismatch}~\citep{ben2007analysis}: different cities, locations, and devices, result in slightly different input representations, and the difference needs to be accounted for to improve overall performance.
Several approaches have been proposed to mitigate the problem, largely drawing from the wide literature of domain adaptation techniques~\citep{ben2007analysis} adapted for the \ac{ASC} problem~\citep{Gharib2018, drossos2019_unsupervised, ren2020caa}, or specifically taking steps to mitigate the effects of city and device~\citep{Chen2019, Komider2019}.

In this work, we adopt a different perspective: we argue that those factors deserve a prominent place in the evaluation of \ac{ASC} systems as they reveal important insights about the behaviour of trained models.
To do that, we adopt the language of recent works in the \ac{ML} \emph{fairness} literature.
In particular, we propose \emph{disaggregated evaluations}, a concept highlighted by \citet{mitchell2019model} as a means to expose the effects that these underlying factors have on system performance.
Disaggregation, which corresponds to breaking down an evaluation to more fine-grained levels of analysis, can be done both in a \emph{unitary} (how performance is affected by each factor independently) and in an \emph{intersectional} way (how performance is affected by combination of factors).
For the task of \ac{ASC}, we consider the three aforementioned factors, namely location, city, and device, as warranting a closer investigation.
This choice is primarily motivated by availability (the existing metadata is already there) and community awareness (past works take them into account).

The rest of this document is organised as follows.
In \cref{sec:fairness}, we formulate our research question by discussing fairness and underspecification for \ac{ASC}.
Our methodological approach, including a description of the data and \ac{DL} architectures used in our experiments, is outlined in \cref{sec:methodology}.
The results and a discussion of our disaggregated evaluations are presented in \cref{sec:experiments}.
Finally, we summarise our findings in \cref{sec:conclusion}.

\section{Fairness and underspecification in \acs{ASC}}
\label{sec:fairness}
The success and increased usage of \ac{ML}, and in particular \ac{DL}, systems in commercial applications has led to rising concerns towards discriminating biases exhibited by \ac{ML} applications, for instance based on race~\citep{Wang_2019_ICCV}.
Especially in the case of \ac{DL}, a lack of interpretability can often be observed~\citep{Burkart2021}, thus posing additional challenges to discover and mitigate said biases.
Even though \ac{ASC} models are not widely considered high risk applications, their increasing usage in smart city~\citep{bello2018sound}, security~\citep{radhakrishnan2005audio}, elderly monitoring~\citep{megret2010immed}, and autonomous driving~\citep{nandwana2016towards} applications means they may soon (or already) be part of critical decision making systems, thus making fairness a critical consideration for those algorithms.

Of the three factors, the recording device is perhaps the most benign; it is hard to justify why an \ac{ASC} system that only works for specific devices should raise ethics concerns, although low-income groups could be excluded if data are only collected with high-end equipment.
On the other hand, city and location (which corresponds e.\,g.\ to specific neighbourhoods) pose potentially bigger problems; a security application should work \emph{equally well} for all citizens irrespective of where they reside, and autonomous driving systems should maintain a standard of performance irrespective of where the vehicle currently is.
There is a already a rich body of work in social sciences discussing inequality across different neighbourhoods on income, health, and other socioeconomic factors~\citep{wen2003poverty}, which an unreliable system may inadvertently exacerbate.
This could have adverse effects against people living in those neighbourhoods, and may disproportionately affect minorities in demographically segregated communities.
Therefore, we anticipate that explicitly communicating disaggregated performance with respect to all three factors would enhance trustability in \ac{ASC} systems used in real-life environmental sensing applications.

Disaggregated evaluations can also be viewed under the perspective of recent research on the \emph{underspecification} of \ac{ML} architectures~\citep{d2020underspecification}, which corresponds to the fact that several architectures yielding similar in-domain performance nevertheless exhibit different behaviour during system deployment.
This undesired property may have negative consequences on the reliability and trustability of \ac{ASC} systems.
For example, if a person using an \ac{ASC} system observed substantially different performance when visiting different neighbourhoods of the same city, they might eventually lose their trust in system performance and stop using it.
As \ac{ASC} architectures increasingly find their way into more real-life applications, the need to address this issue becomes more pressing.
Our evaluation reveals that different architectures yielding almost equivalent performance in standard aggregated evaluations exhibit different behaviour across different sub-populations of the herein examined datasets, thus illustrating that underspecification is also a problem for \ac{ASC} applications.
This shows that disaggregated evaluations can be a useful tool for practitioners that need to select among a pool of candidate models.
% \begin{figure*}[t]
%     \centering
%     \begin{subfigure}{.5\textwidth}
%     \includegraphics[width=\textwidth]{dcase2020_workshop_latex_template/figures/task1-countplot.ps}
%     \end{subfigure}%
%     \begin{subfigure}{.5\textwidth}
%     \includegraphics[width=\textwidth]{dcase2020_workshop_latex_template/figures/task1b-countplot.ps}
%     \end{subfigure}
%     \caption{
%         Countplots showing distribution of samples across different cities and classes for TUT-Urban (left) and TUT-Mobile (right) data sets.
%         For each class, cities are, from left to right: Barcelona, Helsinki, London, Paris, Stockholm, Vienna.
%     }
%     \label{fig:data}
% \end{figure*}

\section{Methodological approach}
\label{sec:methodology}

Our approach consists of the following steps.
First, we train several \ac{DNN} models on the training set of each of the datasets examined here.
Each model is trained for $60$ epochs using \ac{SGD} with a Nesterov momentum of $0.9$, a learning rate of $0.001$, and a batch size of $64$.
For all experiments, we use log Mel spectrograms with $64$-bins as input features, extracted with a window size of $32$\,ms and a hop size of $10$\,ms.
These hyper-parameters were fixed a priori for all models and not optimised during our experiments.
Each model is trained with $5$ random seeds to mitigate the effect of randomness.
%Each model is evaluated in terms of accuracy on the respective evaluation set, and the best model is selected based on that accuracy.

Our experiments are conducted on the TUT Urban Acoustic Scenes 2018 and TUT Urban Acoustic Scenes 2018 Mobile data sets~\citep{Mesaros2018_DCASE}, which will be henceforth referred to as TUT-Urban and TUT-Mobile for brevity.
Both datasets contain data from $10$ acoustic scenes recorded across several locations of $6$ different European cities.
TUT-Urban contains $8640$ stereo samples recorded at $48$\,kHz with a single high-quality recording device (Soundman OKM II Klassik/studio A3), whereas TUT-Mobile additionally contains $720$ samples from each of two additional low-quality recording devices (Samsung Galaxy S7 and iPhone SE).
In the case of TUT-Mobile all data are stored as mono recordings at $16$\,kHz.

\begin{table*}[t]
    \centering
    \caption{
    Aggregated and unitary disaggregated evaluations considering different cities in isolation.
    For the aggregated evaluation, we show accuracy[\%] for all test data for TUT-Urban and TUT-Mobile.
    For the unitary disaggregated evaluations, we show accuracy[\%] on different cities for each architecture, as well as its standard deviation ($\sigma$) over the different cities.
    Results are averaged across $5$ different runs.
    }
    \small
    \label{tab:unitary}
    \begin{tabular}{c|ccccc|ccccc}
        \toprule
        & \multicolumn{5}{c|}{\thead{TUT-Urban}} & \multicolumn{5}{c}{\thead{TUT-Mobile}}\\
         & \thead{FFNN} & \thead{TDNN} & \thead{CNN6} & \thead{CNN10} & \thead{CNN14} & \thead{FFNN} & \thead{TDNN} & \thead{CNN6} & \thead{CNN10} & \thead{CNN14} \\
        \midrule
        \thead{Aggregated} & 52.8 & 57.2 & \textbf{68.7} & 67.7 & 66.3 & 53.1 & 54.7 & \textbf{66.8} & 66.3 & 63.6 \\
        \midrule
        \thead{City} & \multicolumn{10}{c}{\thead{Disaggregated evaluations}}\\
        \midrule
        \thead{Barcelona}  & 52.9 & 61.7 & \textbf{64.8} & 60.9 & 58.9 & 55.9 & 56.4 & \textbf{61.3} & 57.9 & 57.6 \\
        \thead{Helsinki}   & 56.1 & 61.3 & \textbf{70.2} & 67.3 & 63.5 & 50.8 & 57.1 & \textbf{67.9} & 66.7 & 58.3 \\
        \thead{London}     & 51.1 & 61.7 & 71.6 &
        \textbf{74.2} & 71.5 & 52.7 & 59.2 & 70.1 & \textbf{72.0} & 70.8 \\
        \thead{Paris}      & 45.5 & 53.8 & 62.0 & 61.0 & \textbf{62.4} & 45.8 & 54.1 & 60.1 & 59.7 & \textbf{60.8} \\
        \thead{Stockholm}  & 53.0 & 47.4 & \textbf{73.1} & 68.4 & 68.2 & 56.2 & 46.9 & \textbf{72.5} & 68.3 & 67.3 \\
        \thead{Vienna}     & 59.8 & 57.9 & 69.9 & \textbf{74.0} & 73.0 & 58.8 & 54.3 & 68.0 & \textbf{72.4} & 65.7 \\
        \midrule
        $\mathbf{\sigma}$   & 4.4 & 5.2 & \textbf{3.9} & 5.4 & 5.1 & 4.3 & \textbf{3.9} & 4.5 & 5.6 & 4.9 \\
        \bottomrule
    \end{tabular}
\end{table*}
\begin{table*}[t]
    \centering
    \caption{
    Intersectional evaluations considering recording device and city in combination for the TUT-Mobile dataset.
    We show accuracy[\%] for each combination of city and device.
    Cites are Barcelona (B), Helsinki (H), London (L), Paris (P), Stockholm (S), and Vienna (V).
    The best performing architecture value per city and device is marked by boldface.
    Results are averaged across $5$ different runs.
    }
    \label{tab:intersectional}
    \begin{tabular}{c|m{.019\textwidth}m{.019\textwidth}m{.019\textwidth}m{.019\textwidth}m{.019\textwidth}m{.019\textwidth}m{.025\textwidth}|m{.019\textwidth}m{.019\textwidth}m{.019\textwidth}m{.019\textwidth}m{.019\textwidth}m{.019\textwidth}m{.025\textwidth}|m{.019\textwidth}m{.019\textwidth}m{.019\textwidth}m{.019\textwidth}m{.019\textwidth}m{.019\textwidth}m{.019\textwidth}}
        \toprule
         & \multicolumn{7}{c|}{\thead{Device A}} & \multicolumn{7}{c|}{\thead{Device B}} & \multicolumn{7}{c}{\thead{Device C}}\\
        \thead{Model} & \thead{B} & \thead{H} & \thead{L} & \thead{P} & \thead{S} & \thead{V} & $\sigma$ & \thead{B} & \thead{H} & \thead{L} & \thead{P} & \thead{S} & \thead{V} & $\sigma$ & \thead{B} & \thead{H} & \thead{L} & \thead{P} & \thead{S} & \thead{V} & $\sigma$\\
        \midrule
         \thead{FFNN} & 56.1 & 53.1 & 53.0 & 46.9 & 56.5 & 60.5 & \textbf{4.2} & 54.8 & 38.1 & 54.8 & 29.0 & 56.7 & 48.7 & 10.2 & 54.1 & 31.0 & 45.2 & 46.5 & 52.0 & 48.7 & 7.5 \\
        \thead{TDNN} & 57.9 & 60.4 & 62.9 & 57.0 & 47.5 & 56.3 & 4.8 & 44.4 & 38.1 & 26.5 & 34.8 & 53.3 & 41.3 & 8.3 & 46.7 & 30.3 & 34.2 & 32.9 & 31.3 & 43.3 & \textbf{6.2}\\
        \thead{CNN6} & \textbf{61.9} & \textbf{70.2} & 71.2 & \textbf{63.1} & \textbf{73.4} & 69.7 & \textbf{4.2} & \textbf{55.6} & \textbf{58.7} & \textbf{67.7} & 38.1 & 65.3 & \textbf{61.3} & 9.7 & \textbf{57.8} & 44.5 & 56.8 & 39.4 & \textbf{66.0} & 54.7 & 8.8 \\
        \thead{CNN10} & 58.8 & 68.9 & \textbf{73.7} & 62.2 & 69.1 & \textbf{74.2} & 5.6 & 53.3 & 56.1 & 64.5 & 41.9 & \textbf{68.0} & \textbf{61.3} & 8.5 & 49.6 & \textbf{46.5} & 53.5 & 42.6 & 56.7 & \textbf{62.7} & 6.6 \\
        \thead{CNN14} & 58.7 & 60.5 & 71.9 & 62.4 & 68.5 & 67.2 & 4.7 & 48.1 & 47.1 & 64.5 & \textbf{43.2} & 61.3 & 58.0 & \textbf{7.9} & 51.9 & 38.1 & \textbf{60.0} & \textbf{55.5} & 55.3 & 56.0 & 7.0 \\
        \bottomrule
    \end{tabular}
\end{table*}

All models are first evaluated in the standard, aggregated way by computing a single accuracy value, and subsequently assessed using unitary and intersectional evaluations as described below.
We begin with unitary evaluations, where each factor is considered in isolation.
For city and device, where we have only $6$ and $3$ different groups, respectively, we simply report the accuracy for each group.
The location factor is more complicated, as we have $83$ different locations in the test set, thus making it hard to visualise results.
Moreover, whereas for each city and device we have all classes available, each location corresponds to exactly one class, thus making accuracy an inappropriate metric for that evaluation.
To overcome these problems, we compute the $F_1$ score, which is the harmonic mean of precision and recall, for the class corresponding to each location and further normalise the per location $F_1$ score, $F_1^l$, by the overall $F_1$ score for that architecture.
%As models might perform better for some classes vs others, we

Intersectional evaluations are in turn conducted by taking into account two, or more, factors.
Due to space limitations, we only consider results for two pairs of factors: the variation of cities across different devices and the variation across locations in different cities.
For the first case, we report the accuracy for each combination of factors.
For the latter case, we compute the $F_1^l$ score for each location as in the unitary case, but now normalise over the $F_1$ score for each city, $F_1^c$.

As \ac{DL} architectures, we use $5$ standard \ac{DNN} models that belong to different architecture families.
\textbf{\acs{FFNN}}: as the most simple architecture we choose a feed-forward neural network with three hidden layers of decreasing sizes, $300$, $200$, and $100$ units with a \ac{ReLU} activation function. The inputs of of the network are the flattened log Mel spectrograms.
\textbf{\acs{TDNN}}: we further employ a \ac{TDNN} architecture.
First introduced by~\citep{waibel1989phoneme} with the aim of learning temporal relationships, \acp{TDNN} have recently seen great success in the field of speaker identification~\citep{snyder2018x}.
Our \ac{TDNN} architecture is identical to the \ac{DNN} architecture described as the \textit{x-vector system} in~\citep{snyder2018x}.
\textbf{CNN6}, \textbf{CNN10}, \textbf{CNN14}: the final architectures considered in our experiments are \acs{CNN}-based and were recently introduced by \citet{kong2020panns} in the context of audio pattern recognition.
The three architectures have a total of $6$, $10$, and $14$ layers, respectively, excluding pooling layers after convolutional layers, and take Mel-spetrograms as inputs.
The final two layers of each network are fully connected.

\section{Results and discussion}
\label{sec:experiments}
\begin{figure}[t]
    \centering
    \begin{subfigure}{\columnwidth}
    \includegraphics[width=\textwidth]{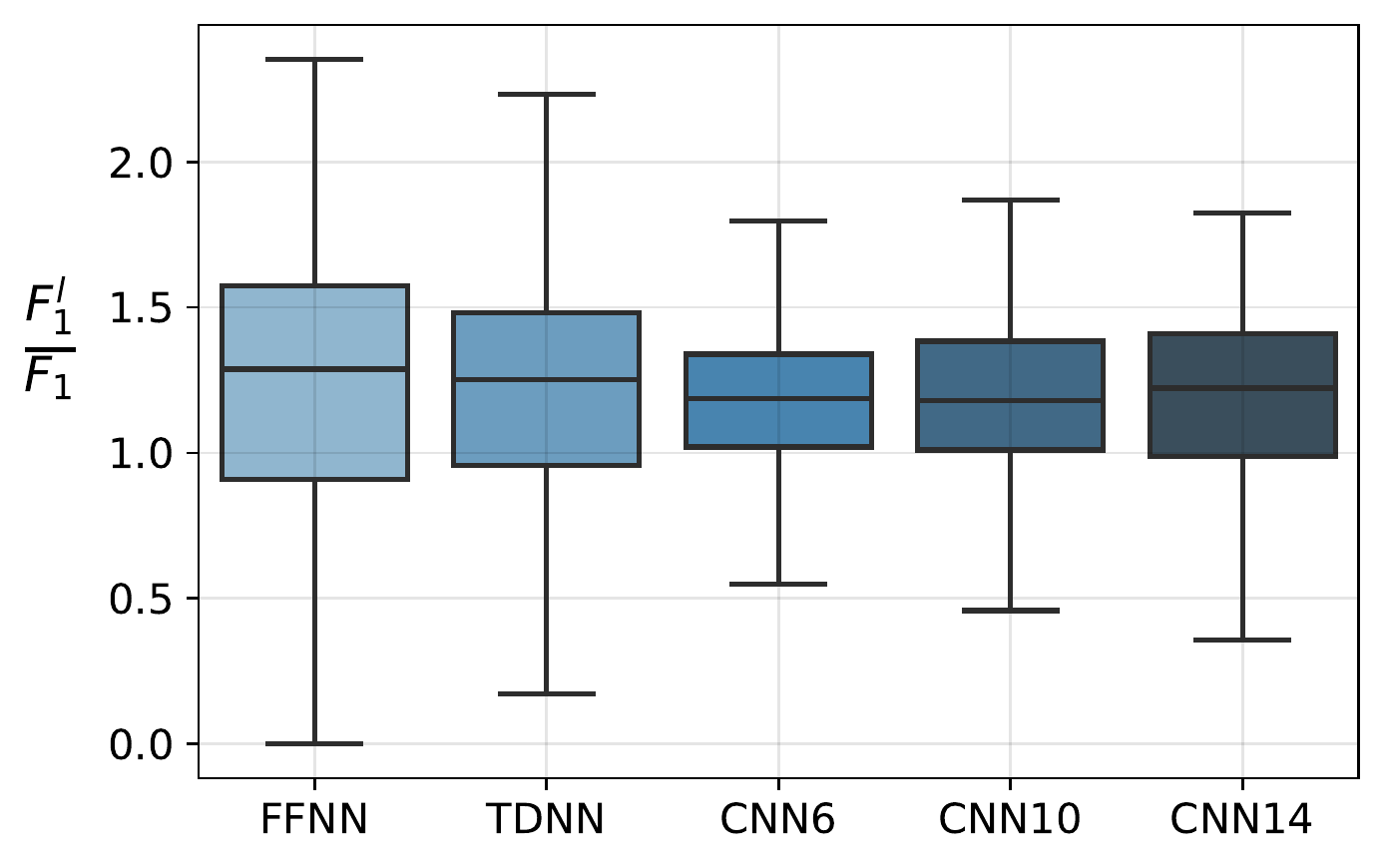}
    \end{subfigure}%
    \caption{
        Distribution of relative $F_1$ score on different locations of TUT-Urban for all architectures.
        Box plots show median and inter-quartile range of relative $F_1$ score.
    }
    \label{fig:unitary}
\end{figure}
\begin{figure}[t]
    \centering
    \begin{subfigure}{\columnwidth}
    \includegraphics[width=\textwidth]{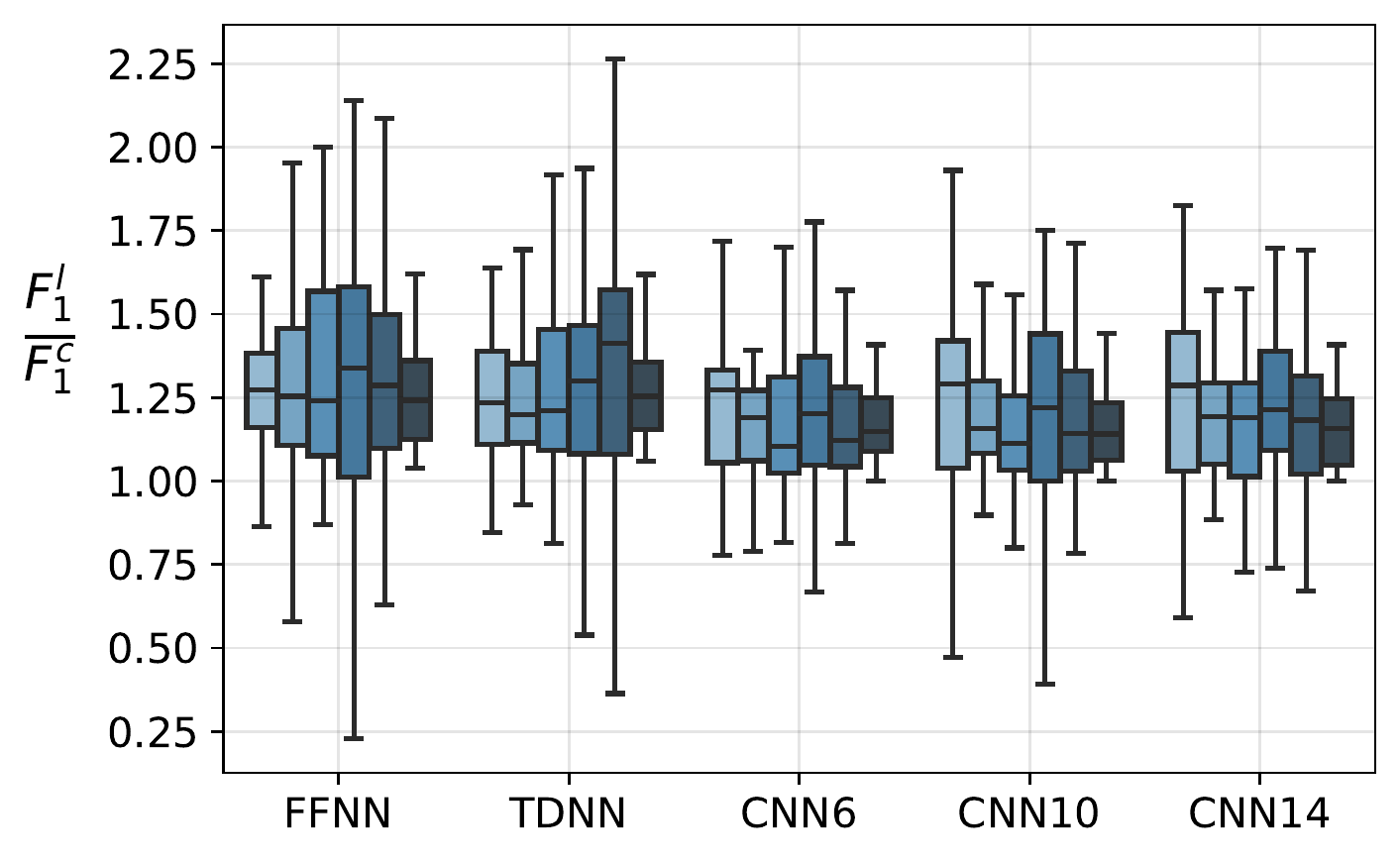}
    \end{subfigure}
    \caption{
    Intersectional analysis of model performance with respect to both city and location for TUT-Urban.
    For each architecture, cities are (from left to right): Barcelona, Helsinki, London, Paris, Stockholm, Vienna.
    Box plots show median and inter-quartile range of relative $F_1$ score with respect to different locations within each city.
    }
    \label{fig:intersectional}
\end{figure}

\begin{figure}
    \centering
    \includegraphics[width=\columnwidth]{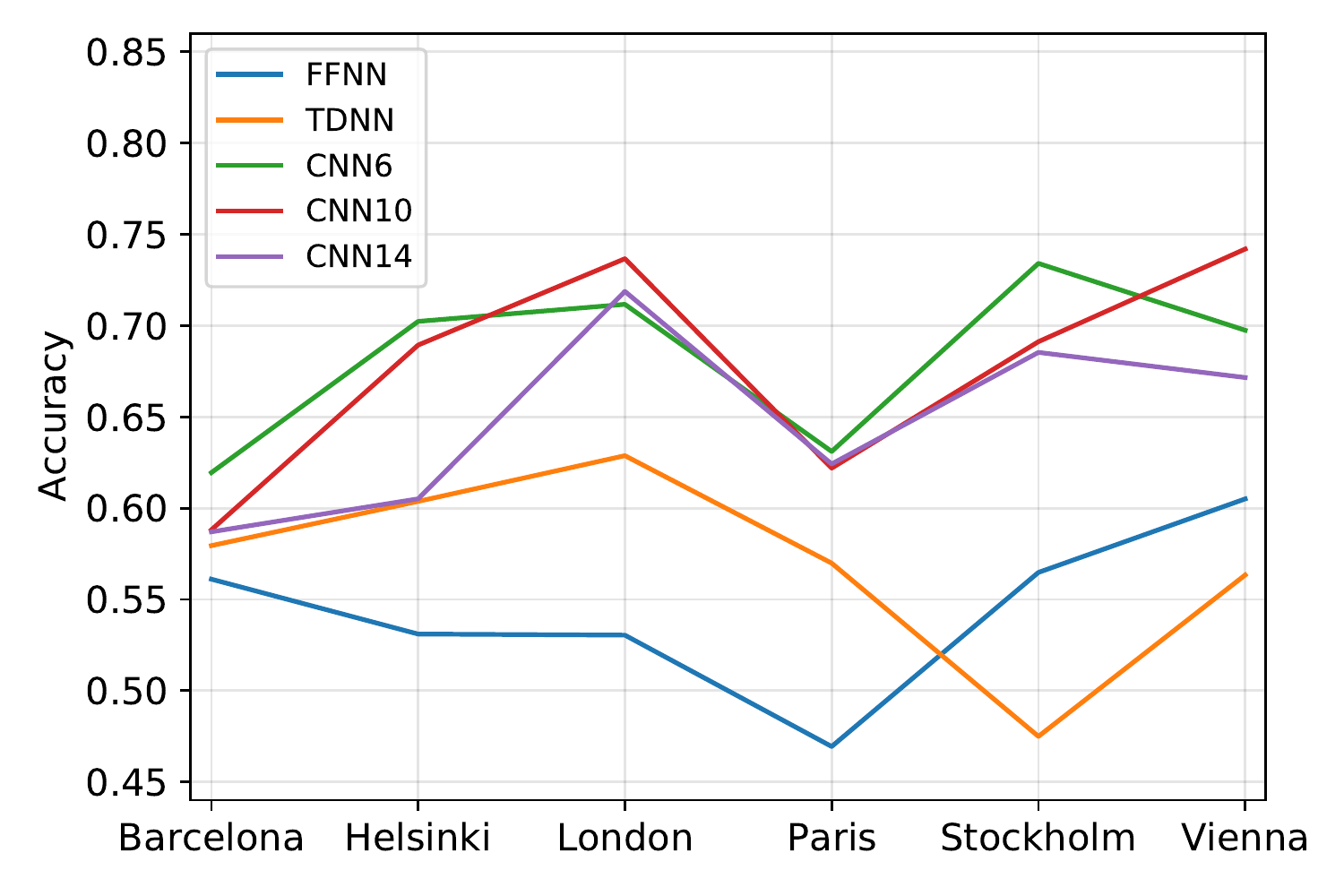}
    \caption{
    Accuracy for each city and architecture on TUT-Mobile.
    }
    \label{fig:underspecification}
\end{figure}

% \begin{figure}
%     \centering
%     \includegraphics[width=\columnwidth]{dcase2020_workshop_latex_template/figures/task1-logmel-city.pdf}
%     \caption{
%     Accuracy for each city and architecture on TUT-Mobile.
%     }
%     \label{fig:my_label}
% \end{figure}

Our unitary evaluation results for different cities are presented in \cref{tab:unitary}, along with the standard aggregated metrics.
We show model accuracy for each factor in isolation, and also report the standard deviation over all factors.
$F_1$ results for different locations in TUT-Urban are shown in \cref{fig:unitary}, where we show box-and-whisker plots of the normalised $F_1$ scores.
We omit unitary results for different devices as they can be inferred from the intersectional results in \cref{tab:intersectional}; as expected, all architectures perform best on the high-quality device A, for which we also have the most data, while doing worse on the lower quality and less populous B and C devices.
Location results on TUT-Mobile are also omitted due to space limitations but exhibit the same trend as those on TUT-Urban.

\cref{tab:unitary} can be read both horizontally, thus emphasising which model works best for a specific factor, and vertically, where we are interested in how a specific model performs across different factors.
Overall, CNN6 is showing the strongest performance, followed by CNN10 and CNN14, with TDNN and FFNN performing substantially worse.
Furthermore, CNN6 exhibits relative stability across both cities and devices.
However, it is not the best choice for all cities; in both datasets, CNN10 is outperforming it for London and Vienna, and CNN14 for Paris, though the latter only marginally.

Of more interest is the vertical interpretation of \cref{tab:unitary}.
We observe that different architectures exhibit a different ordering when it comes to performance per city.
In TUT-Urban for example, different architectures yield their best performance on different cities: FFNN on Vienna, TDNN on Barcelona and London, CNN6 on Stockholm, CNN10 on London and Vienna, and CNN14 on Vienna.
Another interesting case is Stockholm, where CNN6 shows its best performance and TDNN its worst.
Conversely, Vienna, where FFNN, CNN10, and CNN14 show (near-)best performance for TUT-Urban, is showing mediocre results for CNN6 and TDNN.

For TUT-Mobile, these results are better visualised in \cref{fig:underspecification} which shows the range of F1 scores per location for the different cities.
Notable differences exist; TDNN shows worse performance on Stockholm than Paris, whereas all other architectures show the opposite trend.
CNN6 and CNN10, which are almost equivalent in terms of aggregated performance, also exhibit differences, in particular for Stockholm and Vienna.
Interestingly, TDNN and FFNN deviate substantially from the other three architectures, which are more closely clustered together, indicating that models from the same family exhibit more similar behaviour.
These observations illustrate that the inductive biases introduced by each architecture manifest themselves as different behaviours on different strata of each dataset, which is in line with recent research on inductive biases~\citep{ortiz2020neural, ortiz2021neural}.

\cref{fig:unitary} additionally shows that location is a very important factor when it comes to system performance, with some locations exhibiting almost half the aggregated system performance.
Such behaviour is highly undesirable because an \ac{ASC} system deployed across different locations will consistently exhibit subpar performance for some of them, with the risk to equal and fair access to service that this entails.
We note that most locations seem to exhibit better than average performance (the $F_1$ ratio is bigger than 1).
This is caused by the fact that the worst performing locations happen to have more samples, thus having a bigger influence on aggregate performance.

Intersectional results are shown in \cref{tab:intersectional} for the combination of city and device, and in \cref{fig:intersectional} for the combination of city and location.
The differences amongst all cities and all devices were found significant for all architectures using Kruskal-Wallis omnibus H-tests for each factor and architecture, respectively.
% with strong effect sizes
% (FFNN: $.84$, TDNN: $.66$, CNN6: $.81$, CNN10: $.80$, CNN14: $.86$).
% The differences between devices were also significant with strong effect sizes (FFNN: $.64$, TDNN: $.74$, CNN6: $.80$, CNN10: $.88$, CNN14: $.67$).
This shows that, in general, both factors have a large effect on model performance.
In addition, \cref{tab:intersectional} and \cref{fig:intersectional} both show that different architectures exhibit different behaviour on different strata of the two datasets, even though they were trained on identical settings.
Overall, CNN6 is again showing the strongest performance for most, though not all, combinations, followed by CNN10.
In terms of individual factors, Paris is showing the biggest drop in performance when switching from device A to device B for all architectures, indicating that the domain shift introduced by different devices is more adversely impacting this city.

The most interesting case is TDNN, which is showing its best and worst performance on London and Stockholm for device A, respectively, but shows the exact opposite for device B, where the best performance is obtained for Stockholm and the worst for London.
In fact, the performance of TDNN on Stockholm is far better for device B than for device A, even though the latter has far more samples and should thus lead to better performance.

\section{Conclusion}
\label{sec:conclusion}

In this work, we argue for the need of disaggregated unitary and intersectional evaluations for the task of \ac{ASC}.
Our proposed evaluation methodology reveals that several baseline architectures exhibit different behaviour even though they are trained with similar settings.
This illustrates that \ac{ASC} models trained on the examined datasets suffer from the underspecification problem, which heavily impacts the development of reliable and trustworthy systems.
In the future, we intend to further investigate this problem under the perspective of inductive biases introduced by each architecture~\citep{ortiz2021neural}.

Moreover, our work raises interesting questions on the fairness of \ac{ASC} applications.
The architectures examined here exhibit a bias with respect to different cities, locations, and devices.
If these architectures were deployed in a real-world setting, this would translate to non-uniform behaviour over these different factors.
This poses a risk to fair and equitable use of \ac{ML} resources.
We believe this important point needs to be addressed as \ac{ASC} models are being increasingly integrated in intelligent decision making systems.

\section{ACKNOWLEDGMENT}
\label{sec:ack}

Part of the work leading to this publication has received funding from the European Union’s Horizon 2020 research and innovation programme under grant agreement No.\ 957337, project MARVEL.

% -------------------------------------------------------------------------
% Either list references using the bibliography style file IEEEtran.bst
\section{\refname}
\printbibliography[heading=none]

\end{sloppy}
\end{document}